\documentclass[journal]{IEEEtran}

\usepackage{amsmath,amsfonts,amssymb}
\usepackage{algorithmic}
\usepackage{algorithm}
\usepackage{array}
\usepackage[caption=false,font=normalsize,labelfont=sf,textfont=sf]{subfig}
\usepackage{textcomp}
\usepackage{stfloats}
\usepackage{url}
\usepackage{verbatim}

\usepackage{graphicx}
\usepackage{pgfplots}
\usepackage{booktabs}
\usepackage{multirow}
\usepackage{listings}
\usepackage{xcolor}
\usepackage{colortbl}
\pgfplotsset{compat=1.18}

\usepackage{cite}
\usepackage{newtxtext,newtxmath}
\usepackage[T1]{fontenc}

\begin{document}

\title{EVA02-AT: Egocentric Video-Language Understanding with Spatial-Temporal Rotary Positional Embeddings and Symmetric Optimization} 

\author{Xiaoqi Wang,~\IEEEmembership{Student Member,~IEEE,} %
        Yi Wang,~\IEEEmembership{Member,~IEEE,}
        Lap-Pui Chau,~\IEEEmembership{Fellow,~IEEE}

\thanks{The research work was conducted in the JC STEM Lab of Machine Learning and Computer Vision funded by The Hong Kong Jockey Club Charities Trust.}
\thanks{Xiaoqi Wang, Yi Wang, and Lap-Pui Chau are with the Department of Electrical and Electronic Engineering, The Hong Kong Polytechnic University, Hong Kong SAR (e-mail: xiaoqi.wang@connect.polyu.hk; yi-eie.wang@polyu.edu.hk; lap-pui.chau@polyu.edu.hk).}}

\markboth{IEEE Transactions on Image Processing, May 2025}%
{Shell \MakeLowercase{\textit{et al.}}: A Sample Article Using IEEEtran.cls for IEEE Journals}


\maketitle
\IEEEpeerreviewmaketitle

\begin{abstract}
Egocentric video–language understanding demands both high efficiency and accurate spatial-temporal modeling. Existing approaches face three key challenges: 1) Excessive pre-training cost arising from multi-stage pre-training pipelines, 2) Ineffective spatial-temporal encoding due to manually split 3D rotary positional embeddings that hinder feature interactions, and 3) Imprecise learning objectives in soft-label multi-instance retrieval, which neglect negative pair correlations. In this paper, we introduce EVA02-AT, a suite of EVA02-based video-language foundation models tailored to egocentric video understanding tasks. EVA02-AT first efficiently transfers an image-based CLIP model into a unified video encoder via a single-stage pretraining. Second, instead of applying rotary positional embeddings to isolated dimensions, we introduce spatial-temporal rotary positional embeddings along with joint attention, which can effectively encode both spatial and temporal information on the entire hidden dimension. This joint encoding of spatial-temporal features enables the model to learn cross-axis relationships, which are crucial for accurately modeling motion and interaction in videos. Third, focusing on multi-instance video-language retrieval tasks, we introduce the Symmetric Multi-Similarity (SMS) loss and a novel training framework that advances all soft labels for both positive and negative pairs, providing a more precise learning objective. Extensive experiments on Ego4D, EPIC-Kitchens-100, and Charades-Ego under zero-shot and fine-tuning settings demonstrate that EVA02-AT achieves state-of-the-art performance across diverse egocentric video-language tasks with fewer parameters. Models with our SMS loss also show significant performance gains on multi-instance retrieval benchmarks. Our code and models are publicly available at \url{https://github.com/xqwang14/EVA02-AT}.
\end{abstract}

\section{Introduction}
%
%
%
%
\IEEEPARstart{T}{he} research community has witnessed rapid development of egocentric video understanding, driven by improvements in foundation models \cite{I3D,3d_unet, dan_ego, AVION}, pretraining strategies \cite{InternVideo, InternVideo2, egovlpv2}, loss functions \cite{egovlp,RANP}, and data augmentations \cite{lavila}. Despite significant performance gains, the increasing scale of models, prolonged training pipelines, and ever-larger datasets have led to an exponential rise in training costs.

Current state-of-the-art pretraining solutions \cite{InternVideo2,egovideo} generally adopt a pretraining pipeline that involves three stages: 1) capturing the spatial-temporal structure through video reconstruction tasks \cite{videoMAE}, 2) image-text alignment, and 3) video-text alignment via contrastive learning. During the pretraining process, large image and video datasets such as LAION \cite{LAION} and InternVid \cite{internVid}, which contain hundreds of millions of vision-text pairs, make the training process prohibitively expensive.




\begin{figure}[tbp]
    \centering
    \includegraphics[width=1.0\linewidth]{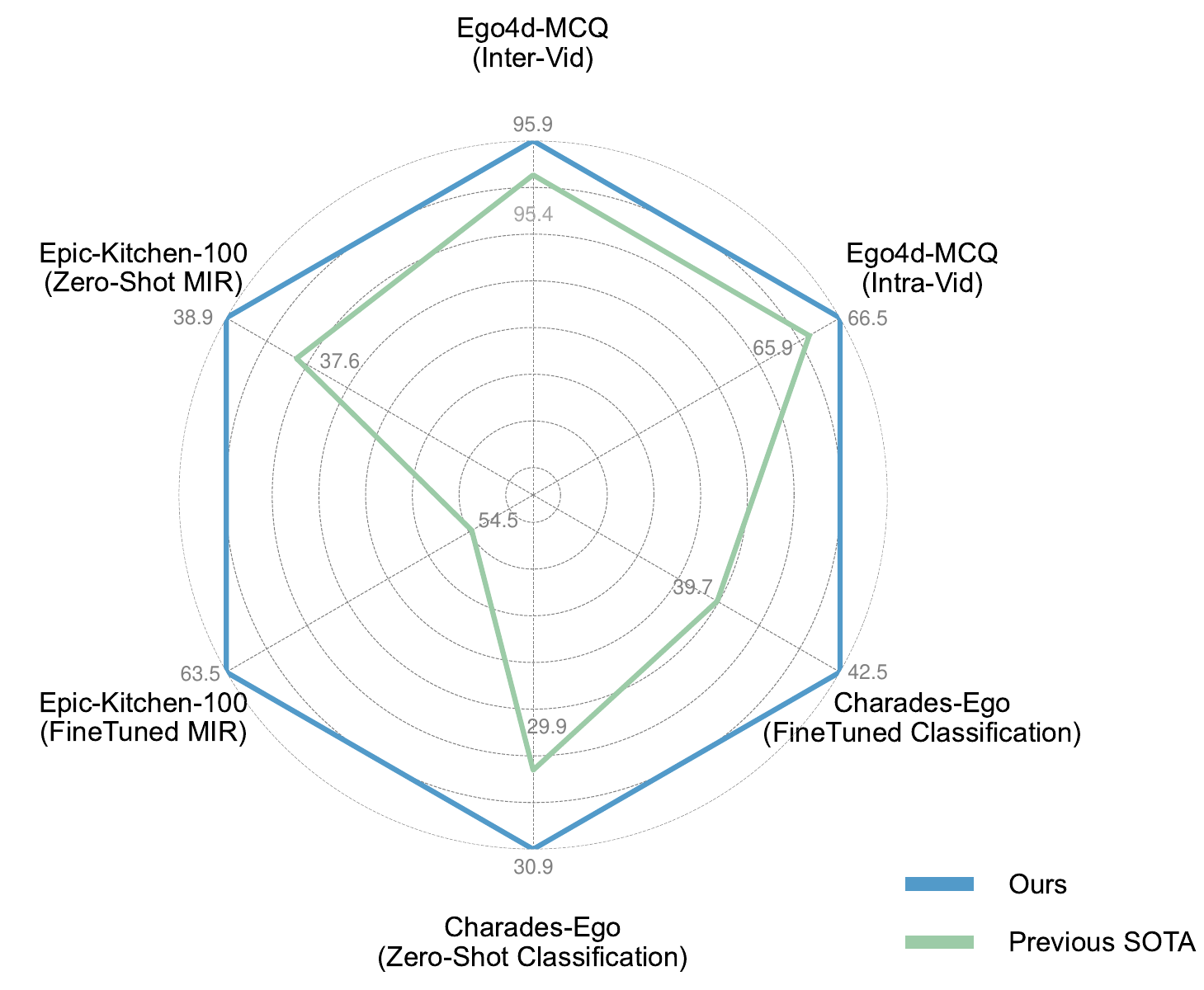}
    \caption{Our EVA02-AT-L model outperforms the previous state-of-the-art methods on three egocentric benchmarks: EgoMCQ, EK-100 MIR, and CharadesEgo in both zero-shot and fine-tune settings by adopting joint attention blocks with integrated spatial-temporal RoPE.}
    \label{fig:introduction}
\end{figure}

Besides the training cost, Rotary Positional Embeddings (RoPE) are now widely used in state-of-the-art vision models \cite{qwen2, eva-clip}. CogvideoX \cite{cogvideoX} first proposes 3D-RoPE, which extends the RoPE to a spatial-temporal approach. Specifically, video tensors in latent space are treated as $(x,y,t)$ coordinates, and CogVideoX applies 1D-RoPE independently at these three coordinates. In practice, the feature dimension is divided into slices of $3/8$, $3/8$, and $1/4$  corresponding to the $x$, $y$, and $t$ coordinates, respectively. Although the effectiveness of this approach has been demonstrated, there are two key issues with the manual division of hidden feature dimensions:

\vspace{-0.5mm}
\begin{itemize}
    \item \textbf{Separation of spatial and temporal embeddings}. The isolation in 3D-RoPE proposed in CogVideoX fails to model cross-axis relationships. Temporal embeddings, which represent motion between frames in video sequences, should ideally reflect changes in the spatial axis over time. In 3D-RoPE, since the dimensions are independent, the time changes $xy + \Delta t$ lack geometric meaning in spatial dimension, preventing the fusion of relative positions across temporal and spatial axes.
    \item \textbf{Uneven dimension division}. Dividing the hidden dimensions of vision transformer architectures into three parts is not always feasible (e.g., 1024 for ViT-L). In the case of 3D-RoPE, the dimensions of the $t$ coordinate are smaller than those of the $x$ and $y$ coordinates, which may be beneficial for spatially sensitive tasks, but reduce the ability to model long video sequences.
\end{itemize}

Moreover, we identified an issue with the current loss functions used in egocentric retrieval tasks. Specifically, EgoVLP \cite{egovlp} introduces the adaptive Multi-Instance Max Margin (MI-MM) loss, which employs a hard mining strategy. This strategy allows the dataloader to select samples where the soft label values exceed a threshold, rather than always selecting the most relevant ones. However, this could lead to negative pairs that are more strongly related to the textual descriptions than the positive pairs, steering the model in the wrong direction. However, simply removing the hard mining strategy would significantly reduce model performance.

To address these issues, we propose \textbf{EVA-02 with spAtial-Temporal attention} (\textbf{EVA02-AT}), a training-efficient solution for egocentric video understanding tasks. The EVA02-AT leverages the image-based pretraining CLIP model of EVA-02 \cite{eva02, eva-clip}, simplifying the pretraining pipeline to a single stage by directly transferring the image-based CLIP model to a video-based one through video-text alignment. To achieve this, we extend the Rotary Positional Embedding (RoPE) to a spatial-temporal approach that is compatible with the original 2D-RoPE. Concretely, RoPE can be treated as a rotation matrix, which is multiplicative, meaning the inner product of two RoPEs equals the sum of their respective positional angles. Therefore, we first generate a 1D-RoPE for the temporal embeddings and a 2D-RoPE for the spatial embeddings, where the dimension of both embeddings corresponds to the whole feature dimension. Then, we conduct an inner product of the temporal and spatial RoPEs to obtain the final representations of our spatial-temporal RoPE. This approach combines the RoPE with learnable temporal and spatial positional embeddings, forming a final positional embedding. Our spatial-temporal RoPE enables each subspace to jointly encode spatiotemporal information, naturally supporting cross-axis relative positions.

To provide a more precise learning objective, we propose the \textbf{Symmetric Multi-Similarity (SMS) loss} to soft label multi-instance retrieval tasks. Inspired by Multi-Similarity loss \cite{ms_loss} and RANP \cite{RANP}, our SMS loss collects not only the correlation values of positive pairs but also the negative pairs, optimizing the model from both sides. Therefore, the SMS loss redefines the relationship between positive and negative pairs and possibly converts certain negative pairs into positive ones under specific conditions, which enables the symmetric optimization of positive and negative pairs. Additionally, we introduce a relaxation factor to SMS loss to avoid the loss from falling into optimizing minor, unimportant samples.

We evaluate our framework on three widely-used egocentric video datasets: Ego4D \cite{ego4d}, EPIC-Kitchen-100(EK-100) \cite{ek100_1,ek100_2}, and Charades-Ego \cite{charades}. The experiment results demonstrate both the effectiveness of our EVA02-AT models and the SMS loss. Our method is able to achieve state-of-the-art performance on these benchmarks in both zero-shot and fine-tuned settings, and the partial results are shown in Fig.\ref{fig:introduction}.

\section{Related Works}

\subsection{Video Foundation Models}

Video foundation models can be grouped by their pretraining pipeline, which is often highly related to their architectural design. The foundation models based on video-text contrastive learning generally extend image–text models by adding temporal modules to capture temporal features. Early work like I3D \cite{I3D} augments spatial 2D-CNNs with an LSTM \cite{LSTM} for temporal feature aggregation. More recent approaches like LaViLa \cite{lavila} and EgoVLP \cite{egovlp, egovlpv2} utilize TSF \cite{TSF} and FiT \cite{FiT} as backbone networks, which add temporal-attention blocks into the ViT backbone, while AVION \cite{AVION} treats each video as a flattened spatial-temporal sequence, processing end-to-end by ViT, greatly reducing overall training costs.

In contrast, models utilizing reconstruction-based pretraining pipelines can learn video representations via self-supervised objectives such as masked video reconstruction \cite{videoMAE, videoMAE2} and next-frame prediction \cite{Flamingo}. This pretraining pipeline trains the model from the beginning, thus facilitating a more flexible architecture. Specifically, Internvideo \cite{InternVideo, InternVideo2} adopts a 3D-CNN in the patchify process to form spatial‐temporal cubes before feeding a ViT, such that the patches contain temporal information, while Flamingo \cite{Flamingo} interleaves cross‐attention layers to jointly encode video and text features.

RoPE \cite{RoPE} has driven recent advances in vision–language models \cite{qwen2, videollama} by providing continuous, unbounded position encoding. However, transferring RoPE to videos remains challenging. As shown in Fig. \ref{fig:rope}, existing solutions like 3D‐RoPE \cite{cogvideoX}, M‐RoPE \cite{qwen2}, and VideoRoPE \cite{video-rope} provide different solutions for video RoPE. 3D-RoPE divides the feature dimension into uneven dimensions and applies three 1D-RoPEs on the entire dimension, so that the three 1D-RoPEs represent $x$-axis, $y$-axis, and $t$-axis individually. VideoRoPE further improves the 3D-RoPE by combining spatial axes, $x$ and $y$, into a uniform 2D-RoPE. However, these methods manually split the embedding dimensions into spatial and temporal parts, such that they preclude a direct transfer of image-based encoders to video domains, and the uneven dimension division may cause a lack of ability to capture temporal information.

\subsection{Loss Functions for Contrastive Learning}

Contrastive learning is a widely adopted paradigm for learning cross-modal representations by aligning paired samples while repelling mismatched ones \cite{c_loss,c_loss_mm,c_loss_mm2}. In video–text pretraining, a common choice is the InfoNCE loss \cite{infoNCE}, which treats video–text pairs as positives and all other pairings within a minibatch as negatives. To better handle noisy alignments, MIL-NCE \cite{mil-nce} relaxes the assumption of perfect correspondence by calculating the summation of multi-instance scores between all positive candidates, while EgoNCE \cite{egovlp} explicitly parses verbs and nouns in captions to weight pairwise affinities according to semantic overlap within each batch.

Beyond batch-wide negatives, several works emphasize the importance of hard negatives or fine-grained similarity metrics. For example, RANP \cite{RANP} mines semantic hard negative pairs and trains in a triplet manner \cite{triplet}, improving the discrimination of closely related but non-matching pairs. Circle loss \cite{circle_loss} and Multi-Similarity (MS) loss \cite{ms_loss} further generalize this idea by weighting each positive and negative pair according to its difficulty, enabling the model to focus more on challenging examples.

Recent advances in soft labeling and adaptive margin strategies have also been shown to improve performance. The adaptive MI-MM loss in EgoVLP \cite{egovlp} incorporates soft labels from EK-100 MIR annotations, achieving a substantial improvement. The relevancy margin loss \cite{relevancy_margin} adds the correlation value on negatives, providing a more accurate learning objective. Inspired by this, we propose the SMS loss, which extends the soft label to both the positive and negative pairs.
\section{Preliminary}

\textbf{Rotary Positional Embedding.} RoPE \cite{RoPE} is known as an effective relative positional embedding approach that has shown extraordinary performance in many state-of-the-art video network architectures \cite{mvd, InternVideo2, cogvideoX}. Originally, the vanilla 1D-RoPE was designed for word embeddings. In transformer-based models that use self-attention mechanisms, RoPE incorporates relative positional information into the attention mechanism. Specifically, the goal of RoPE is to embed the relative position information between the query $\mathbf{x_m}$ at position $m^{th}$ and the key $\mathbf{x_n}$ at position $n_{th}$ within the attention blocks. It should be a function $f(\cdot)$ that satisfies the following condition:

\begin{equation}
\left\langle\boldsymbol{f}_q\left(\boldsymbol{x}_m, m\right), f_k\left(\boldsymbol{x}_n, n\right)\right\rangle=g\left(\boldsymbol{x}_m, \boldsymbol{x}_n, m-n\right),
\end{equation}

where $g(\cdot)$ denotes the real part of the inner product between $f_q\left(x_m, m\right)$ and $f_k\left(x_n, n\right)$. In other words, the inner product between the projected query and key vectors at positions $m$ and $n$ is a function of both the input vectors and their relative position $m-n$. This property indicates that RoPE is a multiplicative positional embedding, meaning the inner product between two RoPE embeddings is equivalent to the subtraction of their corresponding absolute positional embeddings.

\begin{figure}[tbp]
    \centering
    \includegraphics[width=1.0\linewidth]{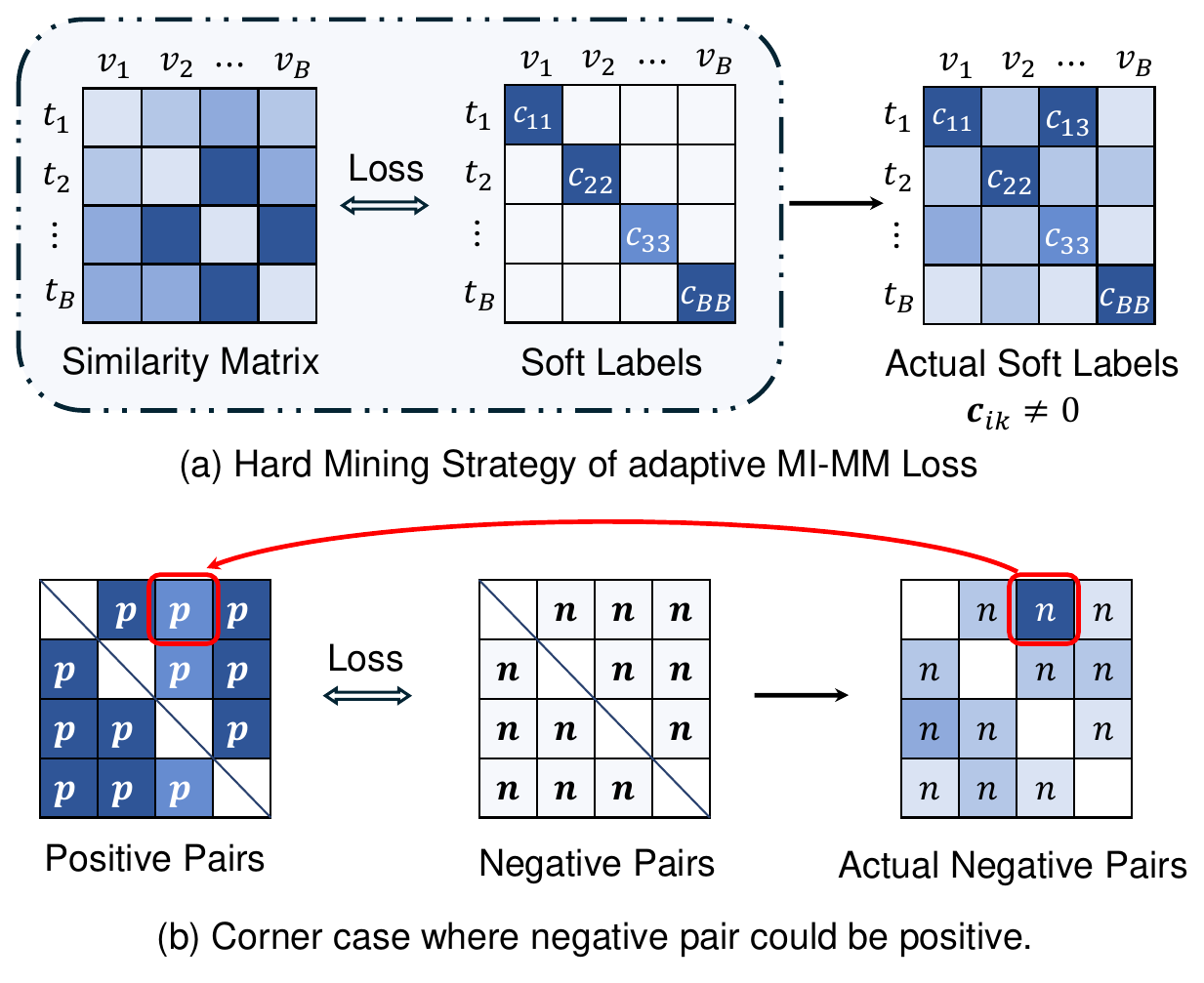}
    \caption{Illustration of the label collection mechanism of adaptive MI-MM loss. Sub-figure (a) indicates that the soft labels collected by the previous dataloader during the training process differ from the actual soft labels, since they only capture correlation values for positive pairs (i.e., the diagonal values). Sub-figure (b)  illustrates a case where negative pairs can have higher correlation values than positive pairs.}
    \label{fig:pre_sms}
\end{figure}

\textbf{Learning objective.} Given a triplet set $\mathcal{D}=\{\mathcal{V}, \mathcal{T}, \mathcal{C}\}$, the objective of the video text retrieval task is to learn a similarity calculation function $S(\cdot)$ that satisfies $S(\mathcal{V}, \mathcal{T}) = \mathcal{C}$. Here, $\mathcal{V} = \{\mathbf{v}_i\}_{i=1}^{\mathcal{N}_v}$ and $\mathcal{T} = \{\mathbf{t}_j\}_{j=1}^{\mathcal{N}_t}$ represent the video and narration sets with $\mathcal{N}_v$ and $\mathcal{N}_t$ samples, respectively. The label $\mathcal{C} = \{\mathbf{c}_{ij} \in \{0,1\} \mid i = 1, 2, \ldots, \mathcal{N}_v, j = 1, 2, \ldots, \mathcal{N}_t\}$ denotes whether a visual-text pair matches, that is, $c_{ij} = 1$ signifies that $(\mathbf{v}_i, \mathbf{t}_j)$ is a corresponding visual-text pair, and vice versa.

In deep metric learning, it is challenging to optimize every sample to its exact position. Alternatively, a general approach is to take advantage of a margin $\gamma$ to separate the positive and negative pairs. Therefore, in the typical visual-to-text retrieval task, the most instinctive learning objective is to ensure that the distance between positive and negative pairs is larger than the margin, which can be formulated as:

\begin{equation}
\begin{aligned}
    \mathcal{O}_{v2t} &:= S(\mathcal{V}, \mathcal{T}_p) - S(\mathcal{V}, \mathcal{T}_n) \geq \mathcal{C}\cdot\gamma,
\end{aligned}
\end{equation}

where $S(\cdot)$ denotes the similarity calculation function, $\mathcal{T}_p$ and $\mathcal{T}_n$ are the matching narrations and mismatching narrations to the corresponding video clips.

Given that $\mathcal{C}$ is the hard label set, where the values can only be 0 or 1, the target distance between the positive and negative pairs for every batch becomes: $(\mathbf{c}_p-\mathbf{c}_n)\gamma=\gamma$. Consider that cosine similarity is used for similarity calculations, where the matrix product of L2-normalized features will represent their similarity, the learning objective becomes:

\begin{equation}
\begin{aligned}
    \mathcal{O}_{v2t} &:= S({\mathbf{v}_i}, {\mathbf{t}_j}) - S({\mathbf{v}_i}, {\mathbf{t}_k}) \geq \gamma\\ 
    &:= \gamma-\mathbf{v}_i^T \mathbf{t}_j + \mathbf{v}_i^T \mathbf{t}_k  \leq 0.
\end{aligned}
\end{equation}

Here, $j$ and $k$ are the samples in positive and negative sets, respectively. Since our task is bidirectional, that is, we need to conduct both the video-to-text and text-to-video retrieval, thus the loss function can be formulated as:

\vspace{0.5mm}
\begin{equation}
    \mathcal{L}=\sum_{(i, j, k) \in \mathcal{N}} \left[\gamma-\mathbf{v}_i^T \mathbf{t}_j+\mathbf{v}_i^T \mathbf{t}_k\right]_+ +\left[\gamma-\mathbf{t}_i^T \mathbf{v}_j+\mathbf{t}_i^T \mathbf{v}_k\right]_+.
\end{equation}

This is a commonly used loss function in the video-text retrieval task, called hinge loss or Multi-Instance Max-Margin (MI-MM) loss \cite{JPoSE}, and $ [\cdot]_+$ denotes the ReLU function here.

Meanwhile, consider a special scenario when soft labels are introduced. In the Epic-Kitchen-100 multi-instance retrieval task, a semantic-based soft label generation method is proposed \cite{relevancy}. Specifically, since narrations are used to describe actions, which can be simplified as the combination of verbs and their corresponding objects(nouns). Consequently, the generation method can be formulated as follows.

\begin{equation}
S_{P o S}\left(y_i, y_j\right)=\sum_{p \in P} \alpha^p \frac{\left|w_i^p \cap w_j^p\right|}{\left|w_i^p \cup w_j^p\right|},
\label{eqn:part-of-speech}
\end{equation}

where $p$ denotes parts of speech, e.g., verb and noun; $\alpha^p$ denotes the weights for every part of speech, commonly $0.5$ for both verb and noun. Therefore, the equation means that the relevancy value, or the soft label values $\mathbf{c}_{ij} \in [0,1]$ between the i-th and j-th narrations equals the IOU of the words in the selected part of the speech. In this scenario, the relevance matrix becomes $\mathcal{C}=\{\mathbf{c}_{ij}\in [0,1]|i=1,2,...,\mathcal{N}_v, j=1,2,...,\mathcal{N}_t\}$. To take advantage of this prior information, the adaptive MI-MM loss \cite{egovlp,egovlp_amm} is proposed, formulated as:

\vspace{0.5mm}
\begin{equation}
    \mathcal{L}=\sum_{(i, j, k) \in \mathcal{N}} [\mathbf{c}_{i j}\gamma-\mathbf{v}_i^T \mathbf{t}_j+\mathbf{v}_i^T \mathbf{t}_k]_+ \\+[\mathbf{c}_{i j}\gamma-\mathbf{t}_i^T \mathbf{v}_j+\mathbf{t}_i^T \mathbf{v}_k]_+.
\end{equation}

The learning objective of adaptive MI-MM Loss is similar to MI-MM Loss, but introduces the relevancy matrix $\mathcal{C}$ to the learning objective. However, the adaptive MI-MM loss only considers the correlations of positive pairs, treating the correlation between video clips and their corresponding negative pairs as 0. As shown in Fig. \ref{fig:pre_sms}(a), the correlation between negative pairs, $c_{ik}$, is not always 0. This makes the learning objective less precise for soft-label-based multi-instance retrieval tasks. Moreover, EgoVLP \cite{egovlp} employs a hard mining strategy that defines the positive set as $i^+=\{j|\mathbf{c}_{ij}\geq 0.1\}$, that is, the partially matched video-text pairs could be treated as positive samples. As illustrated in Fig. \ref{fig:pre_sms}, since adaptive MI-MM loss ignores the correlation values of negative pairs, this can be problematic when $\mathbf{c}_{ij} < \mathbf{c}_{ik}$, leading the learning objective in the opposite direction to the correct one.

\section{The Proposed Method}

\subsection{EVA-02 AT Transformer}

In this subsection, we introduce the design choices in the EVA-02 transformer, including the patchify process, spatial-temporal RoPE embedding, and the theory of joint attention blocks.

\textbf{Patchify.} Inspired by the framework of AVION \cite{AVION}, we integrate a spatial-temporal attention block into a vanilla EVA-02 \cite{ViT, eva02}. For patch embedding, an input video sequence $\mathbf{v} \in \mathbb{R}^{C\times T\times H \times W}$, where $C, T, H, W$ represents channels, number of frames, height, and length, is processed in the spatial domain only. This approach ensures compatibility with the original image encoder, yielding a patchified feature of dimension $\mathbb{R}^{B\times(T\times p^2)\times D}$, where $D =\frac{CHW}{p^2}$.

We introduce two distinct learnable positional embeddings: a temporal positional embedding $P_t \in \mathbb{R}^{T\times D}$ and a spatial positional embedding $P_{xy} \in \mathbb{R}^{{p^2} \times D}$. Each temporal positional embedding is replicated $p^2$ times across the patches of a frame, while each spatial positional embedding is replicated $T$ times to cover all frames. Therefore, the initial representation $z^{(0)}$ after patch embedding is formulated as:

\begin{figure}[tbp]
    \centering
    \includegraphics[width=1.05\linewidth]{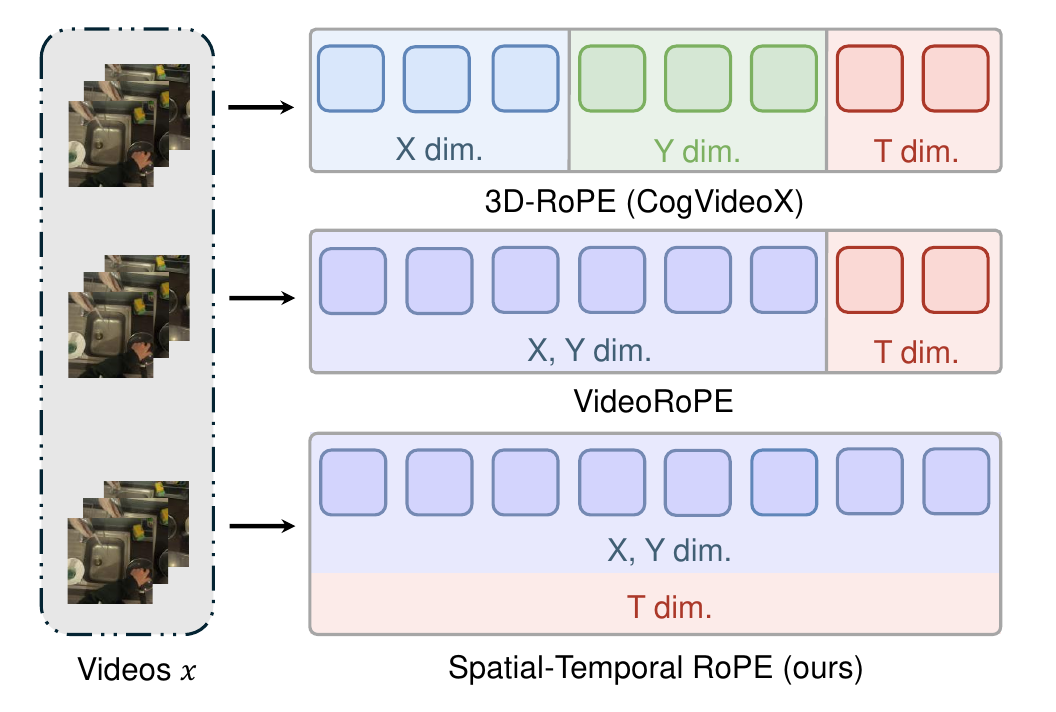}
    \caption{Illustration of different video RoPEs. Our method conducts both spatial and temporal RoPE on the entire feature dimension, forming an integrated spatial-temporal RoPE by leveraging its multiplicative property.}
    \label{fig:rope}
\end{figure}

\begin{figure*}[htbp]
    \centering
    \includegraphics[width=1.0\linewidth]{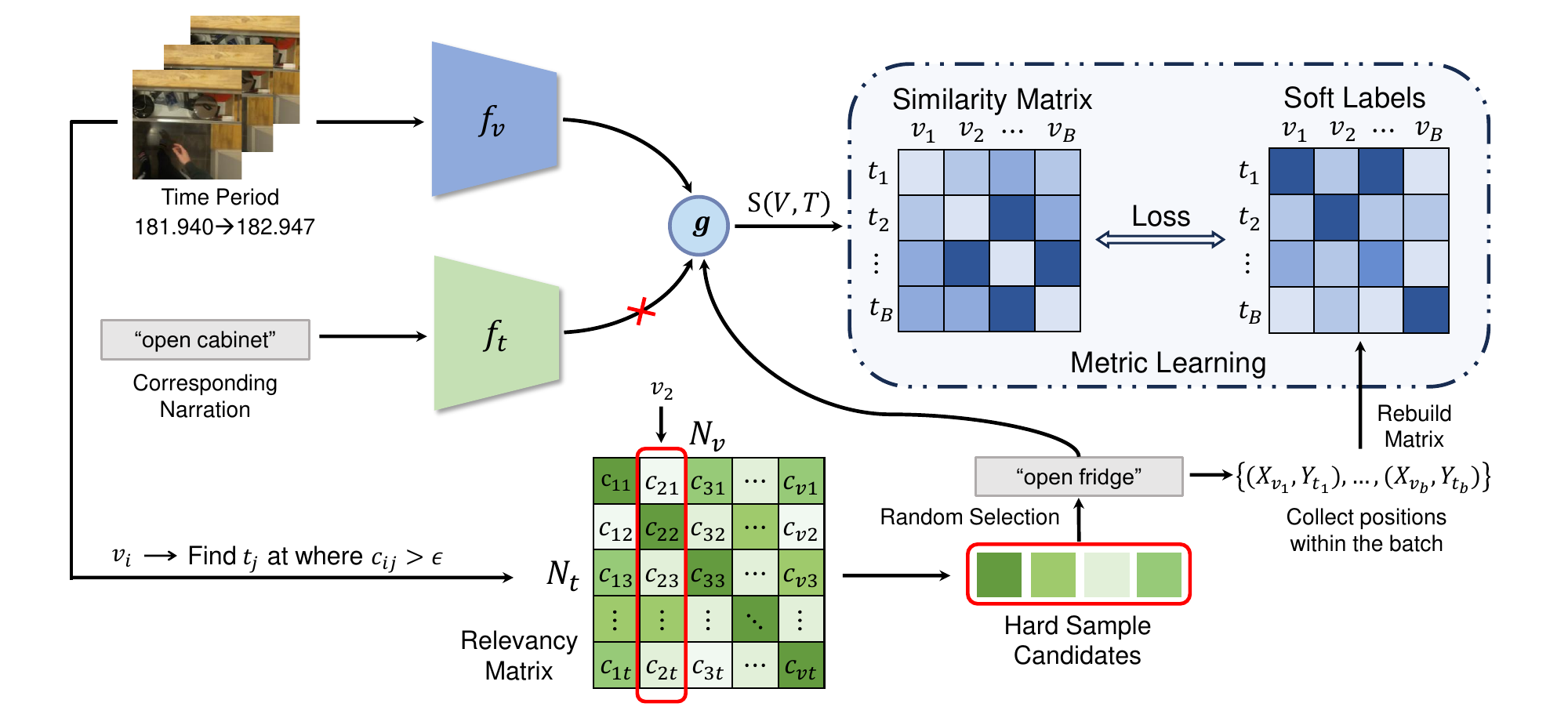}
    \caption{Training framework of EVA02-AT. Given an input video clip ${v}_i$, a hard mining strategy is applied to find a partially matching narration $t_j$ from the pre-build relevancy matrix. Then the dataloader would randomly select a narration as the positive pair to the input video clip from the candidates where the correlation value between ${v}_i$ and narration $\{t_j|j=1,2,...,\mathcal{N_t}\}$ $\mathbf{c}_{ij}$ is greater than a predefined threshold $\epsilon$. Meanwhile, the dataloader will record the serial number of the video clip in pre-build relevancy matrix, thus to rebuild a $B\times B$ correlation matrix during the loss calculation.}
    \label{fig:framework}
\end{figure*}

\begin{equation}
\begin{aligned}
    z^{(0)}&=P^T_{xy}+P^S_{t}+x^{(0)}, \\
    s.t. P^T_{xy}&=\{P^i_{xy}\in\mathbb{R}^{{p^2}\times T \times D} \mid i=1,2,...,t\}, \\
    P^S_{t}&=\{P^j_{xy}\in\mathbb{R}^{{p^2}\times T \times D} \mid j=1,2,...,xy\}.
\end{aligned}
\end{equation}


Here, $P^S_{t}$ and $P^T_{xy}$ denote the final spatial and temporal positional embeddings before the transformer blocks. $x^{(0)}$ denotes the initial feature of the video clip after passing through the first convolutional layer in the patch embedding block.  In this case, we employ a 3D convolution, also known as tube convolution \cite{videoMAE}, with a convolution kernel of $1\times p\times p$. This convolutional operation effectively captures both the spatial and temporal information of the video during the patch embedding phase. The inclusion of temporal dimensions allows the image encoder to act as a video encoder.

\textbf{Joint Spatial-Temporal Attention.} The learnable spatial-temporal positional embedding in EVA02-AT enables the joint spatial-temporal attention. In EVA02-AT, joint attention blocks that process both spatial and temporal information simultaneously are adopted, rather than the divided spatial and temporal attention used in typical video encoders such as Timesformer and Frozen-in-Time \cite{TSF, TBN}.

To cooperate with the joint attention, we need to apply an integrated spatial-temporal RoPE to capture the joint features. Fig. \ref{fig:rope} illustrates how our spatial-temporal RoPE works. Specifically, since the RoPE is a multiplicative positional embedding where the inner product of two RoPEs is equivalent to the addition of rotation angles, to describe a time change in the spatial domain, $xy+\Delta t$, it obeys the following equation:

\begin{equation}
\begin{aligned}
    R_{(xy+\Delta t)}  = R_{xy} \cdot R_{\Delta t}.
\end{aligned}
\end{equation}

Therefore, we initialize a 2D-RoPE $R_{xy} \in \mathbb{R}^{p^2\times D}$ on the spatial domain, where the dimension is evenly divided for height and width, and a 1D temporal RoPE  $R_{t} \in \mathbb{R}^{T\times D}$ on the entire dimension. By calculating the inner product of spatial and temporal RoPE, we obtain an addition of spatial and temporal rotation angles. Similar to the learnable positional embeddings, the spatial RoPE is replicated $T$ times for $T$ frames in the batch, and the temporal RoPE is replicated $p^2$ times for patches in every frame in order to align our 3D-RoPE with the positional embedding. This operation can be expressed as:

\begin{equation}
\begin{aligned}
    R_{(xy+ t)}  &= R^T_{xy} \cdot R^S_{t}, \\
        s.t. R^T_{xy}&=\{R^i_{xy}\in\mathbb{R}^{{p^2}\times T \times D} \mid i=1,2,...,t\}, \\
    R^S_{t}&=\{R^j_{xy}\in\mathbb{R}^{{p^2}\times T \times D} \mid j=1,2,...,xy\}.
 \end{aligned}
\end{equation}

In this way, we thus apply the spatial RoPE and temporal RoPE on the entire dimension instead of manually dividing the dimension into uneven slides. Since we use the standard QK-RoPE, the output of our joint spatial-temporal attention at $k$-th layer can be expressed as:

\begin{equation}
\begin{aligned}
    z^{k} &= SPACE-TIME(z^{k-1}) \\
    \hspace{1mm}
                   &= Attn\left(R_{(xy+t)}W_{q}z^{k-1}, R_{(xy+t)}W_{k}z^{k-1}, W_{v}z^{k-1}\right).\\
\end{aligned}
\end{equation}

The $z^{k-1}$ denotes the output of the $k-1$th layer. In this way, the attention score between query and key becomes a global attention among all the patches in the video clip instead of the spatial attention on a single frame. Meanwhile, the model can still be trained on the basis of an image encoder, which simplifies the pretraining process.



\subsection{Symmetric Multi-Similarity Loss}

As aforementioned, the adaptive MI-MM \cite{egovlp} is not an accurate loss function since the correlation values of negative pairs are not considered. Therefore, to provide a more accurate learning objective, we introduce a novel training framework, which is shown in Fig. \ref{fig:framework}. Building on the hard-mining strategy of EgoVLP \cite{egovlp}, which treats partially matched pairs as positives, the training framework can learn verbs and nouns in natural languages independently. Beyond this, we further refine it by incorporating correlations from both positive and negative samples. 

Specifically, we compute the relevance matrix via Eqn. \ref{eqn:part-of-speech}. During training, the dataloader collects not only matched video–text pairs but also sequences of video $v_i$ and partially matched text $t_j$. For each batch, we reconstruct a $B\times B$ relevance matrix from these sequences, where $B$ represents the batch size. Thus, the relevancy value between arbitrary video and text within the batch can be found in this batch-wise relevancy matrix, so that negative pair entries reflect their true correlation scores rather than defaulting to zero. This enriched matrix serves as the foundation for our SMS loss. 

Given the correlation values for both positive and negative pairs, we aim to create a loss function that can optimize the model from both directions. The Multi-Similarity Loss \cite{ms_loss} provides us a good example and demonstrates its effectiveness on metric learning tasks, which is formulated as:

\begin{equation}
    \begin{aligned}
        \mathcal{L}_{MS}=\frac{1}{\mathcal{N}} \sum_{i=1}^\mathcal{N} & \left\{\frac{1}{\alpha} \log \left[1+\sum_{j \in P_i} e^{-\alpha\left(S_{i j}-\gamma\right)}\right]\right. \\
        + & \left.\frac{1}{\beta} \log \left[1+\sum_{k \in N_i} e^{\beta\left(S_{i k}-\gamma\right)}\right]\right\},
    \end{aligned}
\end{equation}

where $\mathcal{P}_i$ and $\mathcal{N}_i$ refer to the positive and negative sets corresponding to the $i$-th video clip, $\alpha$ and $\beta$ are the scale factors for positive and negative pairs, respectively. To simplify this loss function, we consider a special case when $\alpha, \beta \xrightarrow{}\infty$:

\begin{equation}
\mathcal{L}_{MS}^{'}=\sum_{(i, j, k) \in \mathcal{N}} \left[\gamma - \mathbf{S}_{i j}\right]_+ + \left[\mathbf{S}_{i k} -\gamma \right]_+.
\end{equation}

This reveals that the learning objective for Multi-Similarity Loss is to push positive pairs closer to the margin while pulling negative pairs away from it. This inspires us to define a symmetric loss function for positive and negative pairs. However, as previously illustrated, it is challenging to determine if $\mathbf{t}_j$ and $\mathbf{t}_k$ are relatively more positive to the video clip $\mathbf{v}_i$. Therefore, directly applying Multi-Similarity Loss to this multi-instance retrieval task is still far from satisfactory.

Therefore, we need to define the positive and negative pairs in our training pipeline. Given two narrations $j$ and $k$ corresponding to $i$-th video clip, we formulate the correlation $\mathcal{R}$ between $S_{ij}$ and $S_{ik}$ as follows:

\begin{equation}
    \mathcal{R} =\sum_{(i, j, k) \in \mathcal{N}} \mathbf{c}_{ij} - \mathbf{c}_{ik}.
\end{equation}

In this way, when the correlation factors $\mathcal{R} > 0$, $\mathbf{v}_i$, and $\mathbf{t}_j$ are the relatively more positive pair compared to $\mathbf{v}_i$ and $\mathbf{t}_k$, and vice versa. Following the concept of multi-similarity loss, we extend the adaptive MI-MM loss to a bi-directional and symmetric form:

\begin{equation}
    \mathcal{L}=\sum_{(i, j, k) \in \mathcal{N}}\left\{
\begin{array}{lll}
\left[\mathcal{R}\gamma-S_{i j}+S_{i k} \right]_+&  & {\mathcal{R}>0} \\
\left[-\mathcal{R}\gamma+S_{i j}-S_{i k} \right]_+&  & {\mathcal{R}<0}
\end{array} \right.
\end{equation}

However, a special case happens when $\mathcal{R}=0$, where the distance between $S_{ij}$ and $S_{ik}$ should be optimized to 0. However, in practice, two factors are preventing us from doing so. First, two descriptions with different verbs and nouns could have the same corresponding values. e.g., the current action label is "eat banana", while the partially matched positive pair is "eat apple", and the negative pair is "grab banana". In this case, $\mathbf{c}_{ij}$ and $\mathbf{c}_{ik}$ are the same, but the distance between them should not be optimized to 0.

Meanwhile, we find that the loss at $\mathcal{R}=0$ tends to be the dominant loss since the value of $\mathcal{R}$ is very small. To mitigate this, we introduce a relaxation factor, $\tau$, such that when the Euclidean distance between $S_{i j}$ and $S_{i k}$ is smaller than $\tau$, we cease optimizing this part. This adjustment allows us to maintain the major learning objective, i.e., $\mathcal{O}:=S_{p}-S_{n}>\mathcal{R}\gamma$. Thus, we obtain a symmetric loss regarding the distance between positive and negative pairs:

\begin{equation}
    \mathcal{L}_{SMS}=\sum_{(i, j, k) \in \mathcal{N}}\left\{
\begin{array}{lll}
\left[\mathcal{R}\gamma-S_{i j}+S_{i k} \right]_+&  & {\mathcal{R}>0} \\
\left[-\mathcal{R}\gamma+S_{i j}-S_{i k} \right]_+&  & {\mathcal{R}<0} \\
\left[\Vert S_{i j}-S_{i k} \Vert_{1} - \tau \right]_+&  & {\mathcal{R}=0}
\end{array} \right.
\end{equation}

Here, $S_*$ denotes both the similarity of video-to-text and text-to-video. Additionally, we add a threshold $\lambda$ to constrain the edge conditions, of which the value equals the threshold for selecting positive pairs. Thus, the final loss function becomes:

\begin{equation}
    \mathcal{L}_{SMS}=\sum_{(i, j, k) \in \mathcal{N}}\left\{
\begin{array}{lll}
\left[\mathcal{R}\gamma-S_{i j}+S_{i k} \right]_+&  & {\mathcal{R}\geqslant \lambda} \\
\left[-\mathcal{R}\gamma+S_{i j}-S_{i k} \right]_+&  & {\mathcal{R}\leqslant -\lambda} \\
\left[| S_{i j}-S_{i k} | - \tau \right]_+&  & {\mathcal{|R|}<\lambda}
\end{array} \right.
\end{equation}

Theoretically, the relaxation factor $\tau$ should be less than the minimum value of $\mathcal{C}$ for $\mathcal{C}>0$. This ensures that the optimization process remains effective and balanced across different correlation scenarios. However, in practice, we sometimes need a larger $\tau$ to prevent the model from focusing on similar pairs. Therefore, we obtain the final representation of SMS loss, which optimizes the model symmetrically according to the difference in correlation values.
\section{Experiments}

\subsection{Datasets and Implementation Details}

\textbf{Datasets.} We conduct the experiments on three egocentric datasets: Ego4D, Epic-Kitchens-100 (EK-100), and Charades-Ego. We first pretrain our models on the EgoClip and EgoClip+ versions of the Ego4D dataset, where the EgoClip is proposed by EgoVLP \cite{egovlp}, which contains 3.8 million video-text pairs for training, and the average length for each video clip is about 1 second. The EgoClip+ is proposed by LaViLa \cite{lavila}, which has a 35-million corpus that is augmented by GPT-2 XL \cite{GPT-2}. After pretraining, we evaluate models on the Ego4D Multiple-Choice Questions (EgoMCQ) benchmark. Before fine-tuning, we directly evaluate the pretrained model on EK-100’s multi-instance retrieval  (MIR) challenge and the Charades-Ego action recognition challenge, where the performance will be treated as zero-shot results. After that, we fine-tune the pretrained model on the training set of these two benchmarks, respectively, and evaluate their fine-tuned results.

\textbf{Implementation Details.} We build our EVA02-AT models based on the AVION framework \cite{AVION},  a vanilla ViT-CLIP backbone, and our EVA02-AT-CLIP variants retain the same architecture as EVA02-CLIP except for the modified positional embeddings described in Section 4. We train on 4 $\times$ NVIDIA RTX 6000 Ada GPUs. During both pretraining and fine-tuning, frames are sampled uniformly from each clip at a resolution of $3\times224\times224$, and the dimension of the feature space is set to 256. For our SMS loss, unless specified, we set the SMS-loss margin $\gamma$ to 0.6, and the relaxation factor $\tau$ to 0.1.

\begin{table*}[htbp]
    \centering
    \caption{The main results on EK-100 multi-instance retrieval task. 'PT Dataset' identifies the pretraining dataset, 'Vis Enc.' indicates the visual encoder the models are using. The symbol '*' indicates reproduced results, ``\dag'' denotes that three input modalities are used: RGB, Flow, and Audio. The base-size models are in white rows and large-size models are in gray.}
    \label{tab:main_result_ek100}
    \renewcommand{\arraystretch}{1.25}
    \resizebox{1.0\textwidth}{!}{
    \begin{tabular}{lccc|cccccc}
      \toprule[0.5mm]
       \multirow{2}{*}{\textbf{Methods}} & \multirow{2}{*}{\textbf{PT Dataset}} & \multirow{2}{*}{\textbf{Vis Enc.}} & \multirow{2}{*}{\textbf{\# Frames}} & \multicolumn{3}{c}{\textbf{mAP (\%)}} & \multicolumn{3}{c}{\textbf{nDCG (\%)}} \\
      & & & & V$\rightarrow$T & T$\rightarrow$V & Avg. & V$\rightarrow$T & T$\rightarrow$V & Avg. \\
      \midrule
      \midrule
       MI-MM  & HowTo100M \cite{100m} & S3D \cite{S3D} & 32 & 34.8 & 23.6 & 29.2 & 47.1 & 42.4 & 44.7 \\
       MME \cite{JPoSE} & - & TBN \cite{TBN} & 25\dag & 43.0 & 34.0 & 38.5 & 50.1 & 46.9 & 48.5 \\
       JPoSE \cite{JPoSE} & - & TBN & 25\dag & 49.9 & 38.1 & 44.0 & 55.5 & 51.6 & 53.5 \\
       AVION* \cite{AVION} & WIT \cite{wit} & ViT-B & 16 & 46.8 & 39.9 & 43.4 & 60.0 & 58.0 & 59.0 \\
       AVION + SMS \cite{AVION} & WIT & ViT-B & 16 & \underline{53.8} & \underline{41.4} & \underline{47.6} & \underline{63.2} & \underline{59.2} & \underline{61.2} \\
       EVA02-AT + SMS & Merged2B & EVA02-AT-B & 16 & \textbf{57.6} & \textbf{45.0} & \textbf{51.3} & \textbf{67.1} & \textbf{63.0} & \textbf{65.0} \\
      \rowcolor{gray!20}
       AVION* & WIT & ViT-L & 16 & 51.0 & 44.9 & 47.9 & 64.7 & 62.5 & 63.6 \\
      \rowcolor{gray!20}
       AVION + SMS & WIT & ViT-L & 16 & \underline{60.0} & \underline{47.8} & \underline{53.9} & \underline{68.7} & \underline{64.5} & \underline{66.6} \\
      \rowcolor{gray!20}
       EVA02-AT + SMS & Merged2B & EVA02-AT-L & 16 & \textbf{63.8} & \textbf{52.4} & \textbf{58.1} & \textbf{71.9} & \textbf{67.9} & \textbf{69.9} \\
      \midrule
      \midrule
       EgoVLP \cite{egovlp} & EgoClip & TSF-B \cite{TSF} & 16 & 49.9 & 40.1 & 45.0 & 60.9 & 57.9 & 59.4 \\
       HierVL-SA \cite{hiervl} & EgoClip & FiT-B \cite{FiT} & 16 & - & - & 46.7 & - & - & 61.1 \\
       EgoVLPv2 \cite{egovlpv2} & EgoClip & TSF-B & 16 & - & - & 47.3 & - & - & 61.9 \\
       AVION* & EgoClip  & ViT-B & 16 & 53.3 & 46.6 & 50.0 & 66.3 & 64.0 & 65.1 \\
       AVION + SMS & EgoClip  & ViT-B & 16 & \underline{60.9} & \underline{48.8} & \underline{54.9} & \underline{69.2} & \underline{65.5} & \underline{67.3} \\
       EVA02-AT + SMS & EgoClip  & ViT-B & 16 & \textbf{63.2} & \textbf{51.3} & \textbf{57.3} & \textbf{71.0} & \textbf{67.0} & \textbf{69.0} \\
      \midrule
      \midrule
       LaViLa \cite{lavila} & EgoClip+ & TSF-B & 16 & 55.2 & 45.7 & 50.5 & 66.5 & 63.4 & 65.0 \\
       AVION & EgoClip+ & ViT-B & 16 & 55.9 & 47.8 & 51.8 & 68.2 & 65.4 & 66.8 \\
       AVION + SMS & EgoClip+ & ViT-B & 16 & \underline{62.9} & \underline{51.1} & \underline{57.0}  &\underline{71.2} & \underline{67.3} & \underline{69.2} \\
       EVA02-AT + SMS & EgoClip+ & ViT-B & 16 & \textbf{64.6} & \textbf{53.4} & \textbf{59.0} & \textbf{72.6} & \textbf{69.0} & \textbf{70.8} \\
      \rowcolor{gray!20}
       LaViLa & EgoClip+ & TSF-L & 16 & 54.7 & 47.1 & 50.9 & 68.1 & 64.9 & 66.5 \\
      \rowcolor{gray!20}
       AVION & EgoClip+ & ViT-L & 16 & 57.9 & 51.1 & 54.5 & 70.4 & 67.6 & 69.0 \\
      \rowcolor{gray!20}
       AVION + SMS  & EgoClip+  & ViT-L & 16 & \underline{67.3} & \underline{56.9}  & \underline{62.1} & \underline{74.7} & \underline{71.2} & \underline{73.0}\\
      \rowcolor{gray!20}
       EVA02-AT + SMS & EgoClip+ & EVA02-AT-L & 16 & \textbf{68.7} & \textbf{58.3} & \textbf{63.5} & \textbf{76.1} & \textbf{72.3} & \textbf{74.2} \\
      \bottomrule[0.5mm]
    \end{tabular}
    }
\end{table*}

\begin{table}[tbp]
  \centering
  \caption{Zero-shot and Fine-tuned Results for Video-to-Text Retrieval Task on Charades-Ego dataset and EgoMCQ benchmark. The base-size models are in white rows and large-size models are in gray.}
  \label{tab:charades_ego}
  \renewcommand{\arraystretch}{1.15}
  \resizebox{1.0\linewidth}{!}{
  \begin{tabular}{l|cc|cc}
  \toprule[0.5mm]
    \multirow{2}{*}{\textbf{Method}} & \multicolumn{2}{c}{\textbf{CharadesEgo}} & \multicolumn{2}{c}{\textbf{EgoMCQ Acc.}} \\
    & mAP (ZS) & mAP (FT)   & Inter-vid. & Intra-vid.\\
    \hline
    \multicolumn{5}{l}{\textsc{(EgoClip)}} \\
    \hline
    EgoVLP & 25.0 & 32.1 & 90.6 & 57.2 \\
    HierVL-SA & 26.0 & 33.8  & 90.5 & 52.4 \\
    EgoVLPv2 &  26.2 & 34.1 & 91.0 & 60.9\\
    SViTT-Ego \cite{svitt} & - & - & 92.9 & \underline{65.9} \\
    EVA02-AT  & - & - & \underline{94.8} & 62.0 \\
    \rowcolor{gray!20} EVA02-AT & - & - & \textbf{96.2} & \textbf{66.0} \\
    \hline
    \multicolumn{5}{l}{\textsc{(EgoClip+)}} \\
    \hline
    LaViLa & 26.8 & 33.7 & 93.8 & 59.9\\
    AVION* & \underline{27.4} & 34.8 & 94.5 & 61.4\\
    EVA02-AT &\textbf{27.8} & \underline{36.1}& 95.0 & 63.2\\
    EVA02-AT+SMS & - & \textbf{38.0}& - & -\\
    \rowcolor{gray!20} LaViLa & 28.9 & 36.1 & 94.5 & 63.1 \\
    \rowcolor{gray!20} AVION* & \underline{29.9} & 39.7 & \underline{95.4} & 64.5\\
    \rowcolor{gray!20} EVA02-AT & \textbf{30.9} & \underline{42.2}& \textbf{95.9} & \textbf{66.5} \\
    \rowcolor{gray!20} EVA02-AT+SMS& - & \textbf{42.5}& - & - \\
    \bottomrule
  \end{tabular}
  }
\end{table}

\textbf{Ego4D pretraining.} For Ego4D pretraining, we optimize a bi-directional InfoNCE loss \cite{infoNCE} with the temperature 0.05. We evenly sample 4 frames for each video clip. The batch size for our base-size model is set to 256 per GPU, resulting in a total batch size of 1024, while the batch size is set to 128 for the large model, resulting in a total batch size of 512. We train for five epochs using AdamW \cite{adamw} with a fixed learning rate of $3\times10^{-5}$. The pretraining process takes approximately 40 hours for our base model.

\textbf{EK-100 MIR.} When fine-tuning on the EK-100 dataset, we employ our SMS loss to fine-tune the model pretrained on the Ego4D dataset for 100 epochs. We warm up the learning rate from $10^{-6}$ to a peak of $2\times10^{-5}$ over the first epoch. During fine-tuning, 16 frames are sampled for each video clip, and the batch size is set to 64 for each GPU. Fine-tuning the base model under these settings requires about 20 hours.

\textbf{Charades-Ego Fine-tuning.} The Charades-Ego dataset only contains hard labels, but there could be multiple different hard labels for each video clip. In order to be compatible with the Charades-Ego dataset, we refine our SMS loss as follows:

\begin{equation}
    \mathcal{L}_{SMS}=\sum_{(i, j, k) \in \mathcal{N}}\left\{
\begin{array}{lll}
\left[\mathcal{R}\gamma-S_{i j}+S_{i k} \right]_+&  & {\mathcal{R}=1} \\
\left[| S_{i j}-S_{i k} | - \tau \right]_+&  & {\mathcal{R}=0}
\end{array} \right.
\end{equation}

We fine-tune the model for 10 epochs using the Lamb optimizer, warming up from $10^{-6}$ to $3\times10^{-5}$ in the first epoch. And we sample 16 frames per video clip per GPU. The margin $\gamma$ of SMS loss is set to 0.3.

\begin{table*}[tbp]
    \centering
    \caption{Zero-shot performance of various network architectures on the EK-100 multi-instance retrieval task. The symbol '*' indicates reproduced results. The 'Params (M)' column lists the number of parameters for video encoders, text encoders, and additional blocks (if any), in that order. The base-size models are in white rows and large-size models are in gray.}
    \label{tab:zs_result}
    \renewcommand{\arraystretch}{1.4}
    \resizebox{1.0\textwidth}{!}{
    \begin{tabular}{lccc|cccccc}
      \toprule[0.5mm]
      \multirow{2}{*}{\textbf{Methods}} & \multirow{2}{*}{\textbf{PT Dataset}} & \textbf{Backbone} &\textbf{Params (M)} & \multicolumn{3}{c}{\textbf{Zero-Shot mAP (\%)}} & \multicolumn{3}{c}{\textbf{Zero-Shot nDCG (\%)}} \\
      & & Vis-Text Enc.& Vis-Text Enc.& V$\rightarrow$T & T$\rightarrow$V & Avg. & V$\rightarrow$T & T$\rightarrow$V & Avg. \\
      \midrule
      \midrule
      Random & -         & -        & -              &5.7 & 5.6  & 5.7  & 10.8 & 10.9 & 10.9 \\
      EgoVLP  & EgoClip  & TSF-B + DistillBERT-B \cite{DBERT} &   114 + 66     & 19.4 & 13.9 & 16.6 & 24.1 & 22.0 & 23.1 \\
      HierVL-SA  & EgoClip  & FiT-B + DistillBERT-B &  114 + 66 + 7    & -   & -    & 18.9 & -    & -    & 24.7 \\
      EgoVLPv2& EgoClip  & TSF-B + RoBERT-B \cite{RoBERTA}    &  129 + 138    & -    & -    & 26.7 & -    & -    & 29.1 \\
      AVION & EgoClip  & CLIP-ViT-B      &    86 + 63   & \underline{31.7} & \underline{25.1} & \underline{28.4} & \underline{31.0} & \underline{28.1} & \underline{29.5} \\
      EVA02-AT & EgoClip  & CLIP-EVA02-AT-B &    86 + 63   & \textbf{33.5} & \textbf{26.8} & \textbf{30.2} & \textbf{32.8} & \textbf{29.4} & \textbf{31.1} \\
      \rowcolor{gray!20}
      AVION* & EgoClip  & CLIP-ViT-L      & 304 + 124    & \underline{33.6} & \underline{27.1} & \underline{30.4} & \underline{31.8} & \underline{29.0} & \underline{30.4} \\
      \rowcolor{gray!20}
      EVA02-AT & EgoClip  & CLIP-EVA02-AT-L & 304 + 124     & \textbf{42.1} & \textbf{35.0} & \textbf{38.5} & \textbf{37.2} & \textbf{33.9} & \textbf{35.5} \\
      
      \midrule
      \midrule
      EgoVLP   & EgoClip+  & TSF-B + DistillBERT-B    &   114 + 66   & 26.0 & 20.6 & 23.3 & 28.8 & 27.0 & 27.9 \\
      LaViLa    & EgoClip+   & TSF-B + Text-CLIP-B & 114 + 63 & 35.1 & 26.6 & 30.9 & 33.7 & 30.4 & 32.0 \\
      AVION    & EgoClip+   & CLIP-ViT-B & 86 + 63 & \underline{37.1} & \underline{28.7} & \underline{32.9} & \underline{34.4} & \underline{31.0} & \underline{32.7} \\
      EVA02-AT & EgoClip+ & CLIP-EVA02-AT-B & 86 + 63  & \textbf{38.3} & \textbf{30.3} & \textbf{34.3} & \textbf{36.0} & \textbf{32.2} & \textbf{34.1} \\
      \rowcolor{gray!20}
      LaViLa & EgoClip+ & TSF-L + DistillBERT-B & 404 + 66 & 40.0 & 32.2 & 36.1 & 36.1 & 33.2 & 34.6 \\
      \rowcolor{gray!20}
      AVION & EgoClip+ & CLIP-ViT-L & 304 + 124 & \underline{41.7} & \underline{33.5} & \underline{37.6} & \underline{36.8} & \underline{33.9} & \underline{35.3} \\
      \rowcolor{gray!20}
      EVA02-AT & EgoClip+ & CLIP-EVA02-AT-L & 304 + 124 & \textbf{43.2} & \textbf{34.5} & \textbf{38.9} & \textbf{38.4} & \textbf{33.9} & \textbf{36.2} \\
      \bottomrule[0.5mm]
    \end{tabular}
    }
\end{table*}

\subsection{Compare with State-of-the-Arts}

\textbf{EK100 MIR.} The choice of pretraining data critically affects performance on the EK-100 multi-instance retrieval task. Currently, the state-of-the-art methods are using different pretraining settings, leading to a variety of results. To ensure fair comparisons, we group existing methods by their public pretraining datasets: (a) Image or non-egocentric video dataset; (b) EgoClip \cite{egovlp,ego4d}; and (c) EgoClip with LLM-augmented corpus (EgoClip+) proposed by LaViLa \cite{lavila}. 

Table \ref{tab:main_result_ek100} shows the comparison between the state-of-the-art results and our methods on the EK-100 multi-instance retrieval task, with base-size models in white rows and large-size models in gray. Across all three dataset categories, our models lead both base and large configurations. For base-size models, we improve average mAP by 7.2\% (59.0 vs. 51.8) and average nDCG by 4.0\% over the previous state-of-the-art method, AVION. Scaling to a large-size model, the gain boosts to 9.0\% in average mAP (63.5 vs. 54.5), and 5.2\% (74.2 vs. 69.0) in average nDCG. 

We can also observe from the table that our SMS loss drives much of this improvement. Simply replacing AVION’s MI-MM loss with SMS yields a 7.6\% improvement in average mAP and a 4.0\% improvement in average nDCG. Furthermore, EVA02-AT architectures consistently outperform vanilla ViTs: when training on EgoClip+, our base-size and large-size models improve the performance by 2.0\% and 1.4\% in average mAP, respectively.

\textbf{CharadesEgo Action Recognition.} Table \ref{tab:charades_ego} provides the comparison results on CharadesEgo Video-to-Text action recognition task. Notably, with our SMS loss, our model outperforms the previous state-of-the-art results by 3.2\% on the base model and 2.8\% on the large model in V2T mAP in the fine-tune setup. We also evaluate our EVA02-AT model in a zero-shot setup, and the experiments show that our EVA02-AT outperforms the ViT models by 0.4\% on the base model and 1.0\% on the large model, respectively.

\textbf{EgoMCQ.} We directly evaluate the EgoMCQ performance after pretraining the model on the Ego4D dataset. On EgoMCQ, our base model achieves 95.0\% inter-video accuracy and 63.2\% intra-video accuracy, while our large model achieves 95.9\% inter-video accuracy and 66.5\% intra-video accuracy, which surpasses the previous state-of-the-art results.

\subsection{Ablation Study}

To evaluate the effectiveness of both our EVA02-AT network and the SMS loss function, we conduct the ablation experiments from three aspects: (1) the zero-shot performance across different network architectures; (2) the EVA02-AT model with different temporal positional embedding choices; (3) the fine-tuned performance across different loss functions. 

\begin{table}[htbp]
  \centering
  \caption{Comparison between different temporal embeddings on the zero-shot EK-100 MIR benchmark.}
  \label{tab:pos_emb}
  \renewcommand{\arraystretch}{1.25}
  \begin{tabular}{l|c|cc}
  \toprule[0.5mm]
    \textbf{Backbone} & \textbf{Temporal PE.} & \textbf{mAP} & \textbf{nDCG} \\
    \midrule
    \midrule
    ViT & Learnable & 28.4 & 29.5 \\
    EVA02-AT & Learnable & 28.2 & 29.9 \\
    EVA02-AT &RoPE  & \underline{28.8} & \underline{30.1} \\
    EVA02-AT & Learnable+RoPE & \textbf{30.2} & \textbf{31.1} \\
    \bottomrule
  \end{tabular}
\end{table}

\textbf{Effect of EVA02-AT.} In the zero-shot setting, we evaluate models pretrained on the EgoClip and EgoClip+ datasets, respectively. As Table \ref{tab:zs_result}, our model consistently achieves the state-of-the-art results on both pretraining datasets without increasing the number of parameters. In contrast, backbone models like TSF and FiT, which introduce an external temporal attention block, inevitably increase the model's parameters but fail to provide improved performance. i.e., the EVA02-AT outperforms LaViLa by 2.8\% in average mAP for the large model. Meanwhile, compared with the architectures with joint attention, our model also achieves a better result with the help of spatial-temporal RoPE. i.e., the EVA02-AT beats ViT-B and ViT-L by 1.4\% and 1.3\% in average mAP, respectively.

\begin{figure}[tbp]
    \centering
    \includegraphics[width=1.0\linewidth]{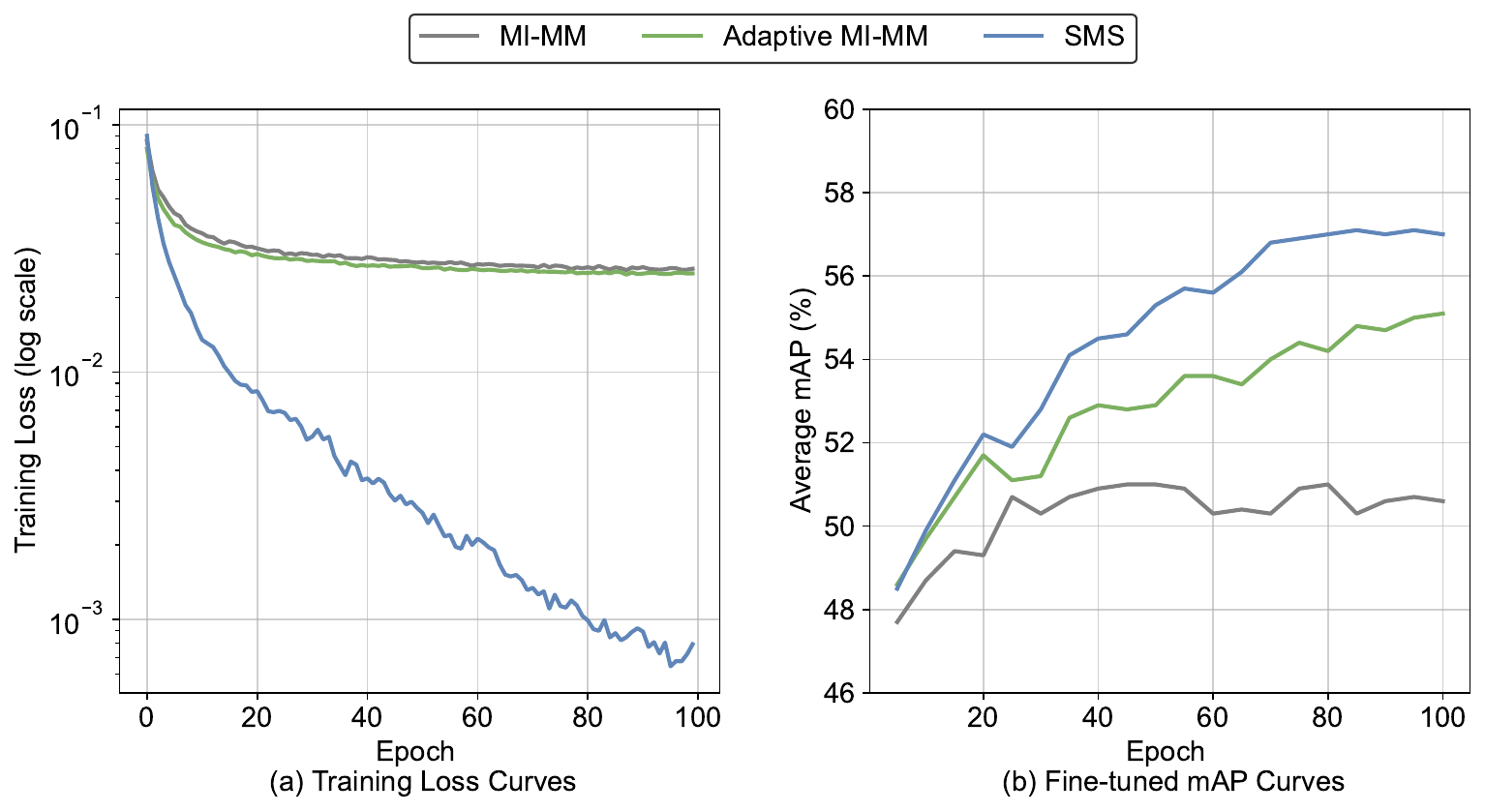}
    \caption{Training curves for different loss functions. Figure (a) shows the loss value during the training process, and Figure (b) shows the validation mAP during the training process on the EK-100 MIR task. By providing an accurate learning objective, SMS decades more sharply than the other two losses.}
    \label{fig:loss_comp}
\end{figure}

\textbf{Effect of 3D-RoPE.} In table \ref{tab:pos_emb}, we change the temporal positional embedding to (a) the learnable positional embedding, (b) 1D-RoPE embedding, and (c) learnable positional embedding with RoPE embedding. From the table, we can find that changing the temporal positional embedding will not influence the performance significantly, but (c) still outperforms all the other settings. Concretely, compared to the learnable temporal positional embedding, RoPE improves the model by 2.0\% in average mAP. And compared to the case that only uses temporal RoPE, a learnable positional embedding can provide a 1.4\$ gain in average mAP. Additionally, the experiment suggests that preserving the model's extrapolation ability, i.e., using RoPE as the only temporal positional embedding, does not lead to a noticeable performance drop.

\begin{table}[tbp]
    \centering
    \caption{The performance comparison of different loss functions pretrained on the EgoClip+ dataset and fine-tuned on the EK-100 multi-instance retrieval task.}
    \resizebox{1.0\linewidth}{!}{
    \begin{tabular}{cc|cccccc}
      \toprule[0.5mm]
      &\multirow{2}{*}{\textbf{Methods}}& \multicolumn{3}{c}{\textbf{ mAP (\%)}} & \multicolumn{3}{c}{\textbf{ nDCG (\%)}}\\
      & & V$\rightarrow$ T & T$\rightarrow$ V & Avg. & V$\rightarrow$ T & T$\rightarrow$ V & Avg. \\
      \midrule
      \midrule
      \multirow{5}{*}{\rotatebox{90}{\textit{ViT-B-16}}}
      &AVION & 55.9 & 47.8 & 51.8 & 68.2 & 65.4 & 66.8 \\
      &MI-MM*  & 55.5  & 48.8 & 52.1 & 68.4 & 66.3 & 67.3  \\
      &Adaptive MI-MM & 60.5  & \underline{49.6} & 55.1 & 69.7 & \underline{66.5} & 68.1  \\
      &SMS w/o $\tau$ & \underline{62.2}  & 48.1 & \underline{55.2} & \underline{70.8} & \underline{66.5} & \underline{68.6}  \\
      &SMS (ours) & \textbf{62.9} & \textbf{51.1} & \textbf{57.0}  &\textbf{71.2} & \textbf{67.3} & \textbf{69.2} \\
      \midrule
      \midrule
      \multirow{4}{*}{\rotatebox{90}{\textit{ViT-L-14}}}
      &AVION & 57.9 & 51.1 & 54.5 & 70.4 & 67.6 & 69.0 \\
      &MI-MM* & 58.7  & 52.7 & 55.7 & 71.9 & 69.4 & 70.6  \\
      &Adaptive MI-MM & \underline{65.0}  & \underline{54.6} & \underline{59.8} & \underline{73.3} & \underline{70.0} & \underline{71.6}  \\
      &SMS (ours) & \textbf{67.3} & \textbf{56.9}  & \textbf{62.1} & \textbf{74.7} & \textbf{71.2} & \textbf{73.0}\\
      \cmidrule{1-8}
      &w/ EVA02-AT-B & 64.6 & 53.4 & 59.0 & 72.6 & 69.0 & 70.8 \\
      &$\Delta$ - ViT-B-16 & +1.7 & +2.3 & +2.0 & +1.4 & +1.7 & +1.5 \\
      \rowcolor{gray!20}
      &w/ EVA02-AT-L & 68.7 & 58.3 & 63.5 & 76.1 & 72.3 & 74.2 \\
      \rowcolor{gray!20}
      &$\Delta$ - ViT-L-14  & +1.4 & +1.4 & +1.4 & +1.4 & +1.1 & +1.2 \\
      \bottomrule[0.5mm]
    \end{tabular}
    }
    
    \label{tab:result}
\end{table}

\textbf{Effect of SMS Loss.} To verify the effectiveness and robustness of our SMS loss, we conduct an ablation study on both our ViT-B-based and ViT-L-based models. All experiments across different loss functions are conducted under the same learning rate and optimizer settings. We choose the best-performing hyperparameters for each loss function, i.e., a margin of 0.2 for the MI-MM loss and 0.4 for the adaptive MI-MM loss. The experiment results are presented in Table \ref{tab:result}.

\begin{figure}[t]
    \centering
    \includegraphics[width=1.0\linewidth]{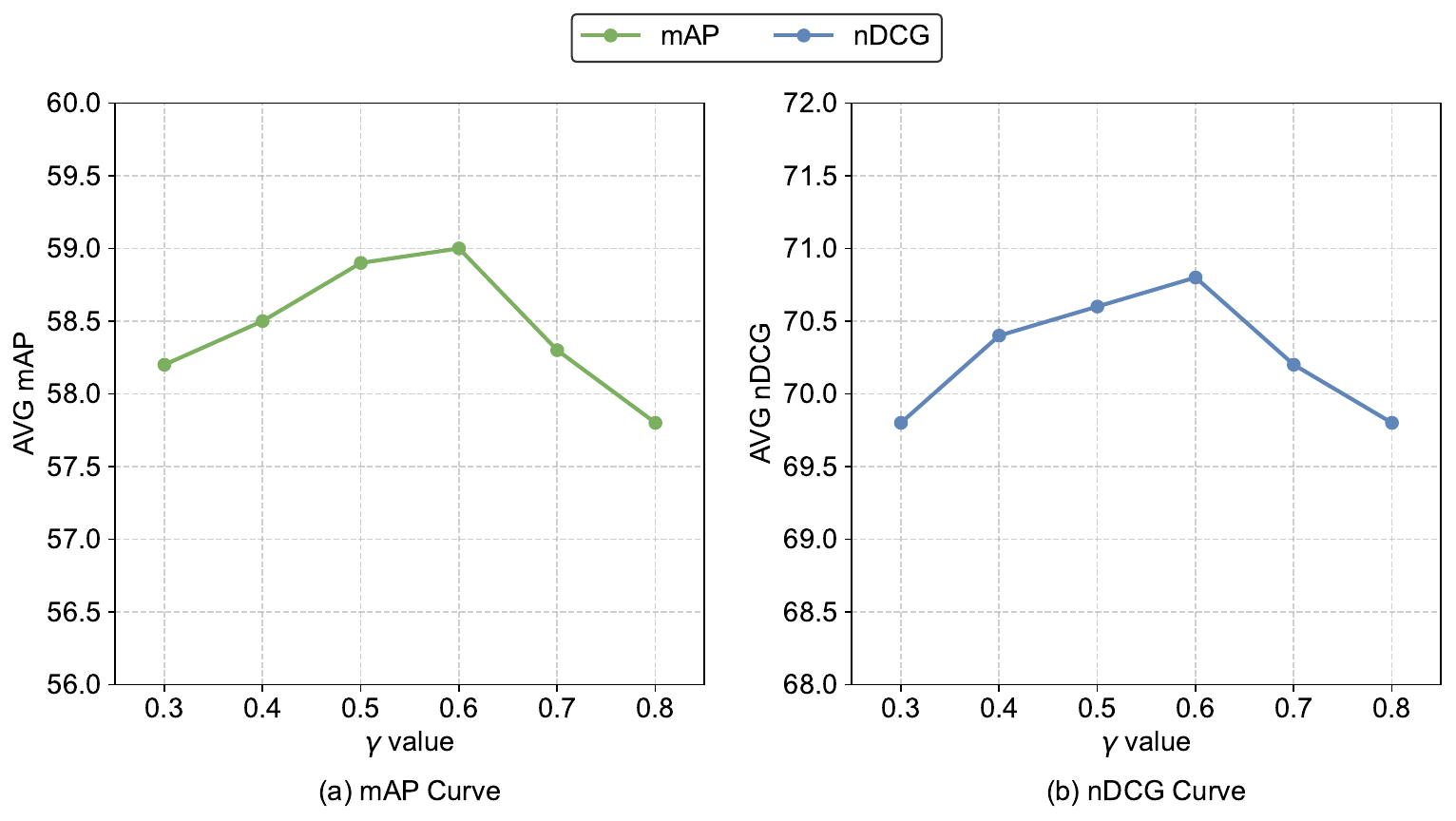}
    \caption{Training curves for different hyper-parameter choices in SMS loss. Figures (a) and (b) show the average mAP and nDCG performances when $\gamma$ changes. }
    \label{fig:gamma}
\end{figure}

For both the ViT-B-based and ViT-L-based models, our SMS loss demonstrates superior performance compared to its counterparts. Specifically, for the ViT-B-based model, our SMS loss improves the average mAP by 1.9\% and the average nDCG by 2.4\% compared to the adaptive MI-MM loss. Similarly, for the ViT-L-based model, our SMS loss also improves the model in average mAP by 2.3\% and in average nDCG by 2.4\%.

We also conducted an experiment using a ViT-B-based model with $\tau = 0$ to evaluate the impact of the relaxation factor. This factor helps prevent over-optimization when the correlation values between positive and negative samples are similar. Our results show a performance drop of 1.8\% in average mAP and 0.8\% in average nDCG compared to the case where $\tau = 0.1$.  This demonstrates the crucial role of the relaxation factor in ensuring optimal model performance.

We next examine the impact of the SMS loss hyperparameter, $\gamma$.  Fig. \ref{fig:gamma} plots average mAP as $\gamma$ varies from 0.3 to 0.8.  Our results show that the model achieves its highest mAP at $\gamma = 0.6$.  Moreover, performance remains stable across the entire range, since the mAP difference between the best and worst settings is only 1.2\%.

The training curves for different loss functions are presented in Fig. \ref{fig:loss_comp}. Notably, a performance gap emerges as early as 20 epochs, with the SMS loss continuing to decrease exponentially, while the other two loss functions show slower declines over the next 80 epochs. Although the absolute value of SMS loss is naturally lower than that of MI-MM and adaptive MI-MM losses, the results highlight that an accurate learning objective significantly helps the fine-tuning process.
\section{Conclusion}
This paper proposes the EVA02-AT suite, a strong and training-efficient video-text CLIP model. Instead of divided spatial and temporal attention blocks used in typical video encoders, we adopt a joint attention block along with an integrated Spatial-Temporal RoPE, which conducts both spatial and temporal RoPE on the entire feature dimension. This approach enables global attention across all the patches within video clips, and avoids an uneven manual division of the feature dimension. The EVA02-AT can be trained on the basis of the image CLIP model, achieves the generalized egocentric video representations without increasing the number of parameters, and surpasses all the previous state-of-the-art results on major egocentric benchmarks. Additionally, we propose the SMS loss, which significantly advances the state-of-the-art performance on the EK-100 MIR task.

\bibliographystyle{IEEEtran}
\bibliography{main.bib}

 





\end{document}